\documentclass{article}
\usepackage{stywhispers}

\usepackage{graphicx}
\usepackage{amsfonts}
\usepackage{mathtools}
\usepackage{amsmath}

\usepackage{cite}

\usepackage{silence}
\graphicspath{{images/}}

\usepackage{pgfplots}
\pgfplotsset{
    compat=newest,
    grid=both,
    minor tick num=4,
    grid style={line width=.1pt, draw=gray!10},
    major grid style={line width=.2pt,draw=gray!50},
    every axis plot/.append style={line width=0.8pt},
}

\usepackage{tikz}
\usetikzlibrary{arrows,positioning,calc,shapes.geometric,patterns,patterns.meta,fit}

\tikzstyle{block}=[
    draw,
    minimum width={width("Global Pooling")+7pt},
    minimum height={height("Sigmoid GLobal Pooling FC-MLP")+7pt}
]
\tikzstyle{block2}=[
    draw,
    minimum width={width("ConvT-120-3x3 $\downarrow^2$ CNorm SE ReLU")+7pt},
    minimum height={height("ConvT-120-3x3 $\downarrow^2$ CNorm SE ReLU")+7pt}
]
\tikzstyle{block3}=[
    draw,
    minimum width={width("Conv-960-3x3 CNorm ReLU")+7pt},
    minimum height={height("Conv-960-3x3 CNorm ReLU")+7pt}
]

\usepackage[labelformat=simple]{subcaption}

\usepackage{array}
\newcolumntype{?}{!{\vrule width 1.2pt}}

\usepackage[hidelinks]{hyperref}

\title{Generative Adversarial Networks for Spatio-Spectral Compression of Hyperspectral Images}

\name{
    Martin~Hermann~Paul~Fuchs$^1$, Akshara~Preethy~Byju$^2$, Alisa~Walda$^1$, Behnood Rasti$^{1,3}$, Beg\"um~Demir$^{1,3}$%
    \thanks{This work is funded by the European Research Council (ERC) through the ERC-2017-STG BigEarth Project under Grant 759764. The Authors would like to thank Nimisha Thekke Madam for the joint discussions during the initial phase of this study.}
}
\address{%
    $^1$Faculty of Electrical Engineering and Computer Science, Technische Universit\"at Berlin, Germany.\\%
    $^2$Department of Computer Science and Engineering, Amrita School of Computing, Amrita Vishwa\\Vidyapeetham, Amritapuri, India.\\%
    $^3$BIFOLD - Berlin Institute for the Foundations of Learning and Data, Germany.%
}

\begin{document}

\maketitle

\begin{abstract}
Deep learning-based hyperspectral image (HSI) compression has recently attracted great attention in remote sensing due to the growth of hyperspectral data archives. Most of the existing models achieve either spectral or spatial compression and do not jointly consider the spatio-spectral redundancies present in HSIs. To address this problem, in this paper, we propose High Fidelity Compression (HiFiC)-based models for spatio-spectral compression of HSIs. In detail, we introduce two new models: i) HiFiC using  Squeeze and Excitation (SE) blocks (denoted as HiFiC$_{SE}$); and ii) HiFiC with 3D convolutions (denoted as HiFiC$_{3D}$) in the framework of compression of HSIs. We analyze the effectiveness of HiFiC$_{SE}$ and HiFiC$_{3D}$ in compressing the spatio-spectral redundancies with channel attention and inter-dependency analysis. Experimental results show the efficacy of the proposed models in performing spatio-spectral compression, while reconstructing images at reduced bitrates with higher reconstruction quality. The code of the proposed models is publicly available at \url{https://git.tu-berlin.de/rsim/HSI-SSC}.
\end{abstract}


\section{Introduction}
\label{sec:intro}

Hyperspectral sensors acquire images characterized by a very high spectral resolution that results in hundreds of observation channels. Dense spectral information provided in hyperspectral images (HSIs) leads to a very high capability for the identification and discrimination of the materials in a given scene. 
However, the huge amount of data results in a heavy burden for the storage and transmission of HSIs \cite{Atli}. Accordingly, the development of efficient compression techniques that encode the data into fewer bits with minimal loss of information is a growing research interest in remote sensing.

HSI compression methods can be grouped into two categories: i) traditional approaches; and ii) learning-based approaches. Traditional approaches mostly rely on transform coding. In \cite{du2007hyperspectral}, HSIs are compressed by applying the JPEG 2000 \cite{rabbani2002book} algorithm in combination with a principal component analysis for spectral decorrelation. Lim et. al \cite{lim2001compression} apply a three-dimensional wavelet transform to compress the HSIs. Abousleman et al. \cite{abousleman1995compression} use differential pulse code modulation to spectrally decorrelate the data, while a 2D discrete cosine transform coding scheme is used for spatial decorrelation. Deep learning-based hyperspectral image compression techniques use convolutional autoencoders (CAEs) \cite{kuester20211d, kuester2022transferability, la2022hyperspectral, chong2021end} that reduce the dimensionality of the latent space by sequentially applying convolutions (convs) and downsampling operations. In \cite{kuester20211d, kuester2022transferability}, a 1D-CAE  model that only compresses the spectral content by stacking multiple 1D convs combined with two pooling layers and LeakyReLU activation functions is introduced. In \cite{la2022hyperspectral}, a novel spectral signal compressor network (SSCNet), which incorporates spatial compression by utilizing 2D convs, PReLUs and 2D max poolings for encoding and strided 2D transposed convs for decoding, is presented. In \cite{chong2021end}, a 3D-CAE that combines spatial and spectral compression is introduced. That network is built by strided 3D convs, LeakyReLU, 3D batch normalization, residual blocks, and upsampling layers in the decoder. However, due to the fixed compression ratio of CAEs and their relatively high bitrates, there is a need for developing more effective approaches.
To address the limitations of CAEs, generative adversarial network (GAN)-based image compression models \cite{mentzer2020highfidelity,wu2020ganbased} have been proposed in the computer vision (CV) community. GANs are capable of producing visually convincing results that are perceptually similar to the input image, and thus GAN-based compression shows improved performance in spatial compression even at extremely low bitrates keeping intact the perceptual quality of the reconstructed image. However, the applicability of GANs in spatio-spectral compression of HSIs has not been studied yet.

In this paper, we propose GAN-based High Fidelity Compression (HiFiC)-based models \cite{mentzer2020highfidelity}. HiFiC consists of 2D convs, and thus treats each band separately to perform band-by-band spatial compression. To effectively compress the spectral and spatial information, we propose two new models: i) HiFiC$_{SE}$ using Squeeze and Excitation (SE) blocks \cite{squeeze}; and ii) HiFiC$_{3D}$ using 3D convs. The SE blocks reduce feature redundancies using channel attention, while 3D convs use 3D kernels to learn feature dependencies across the channels. 
Experimental results show that the proposed models achieve low bitrates with good perceptual reconstruction. 
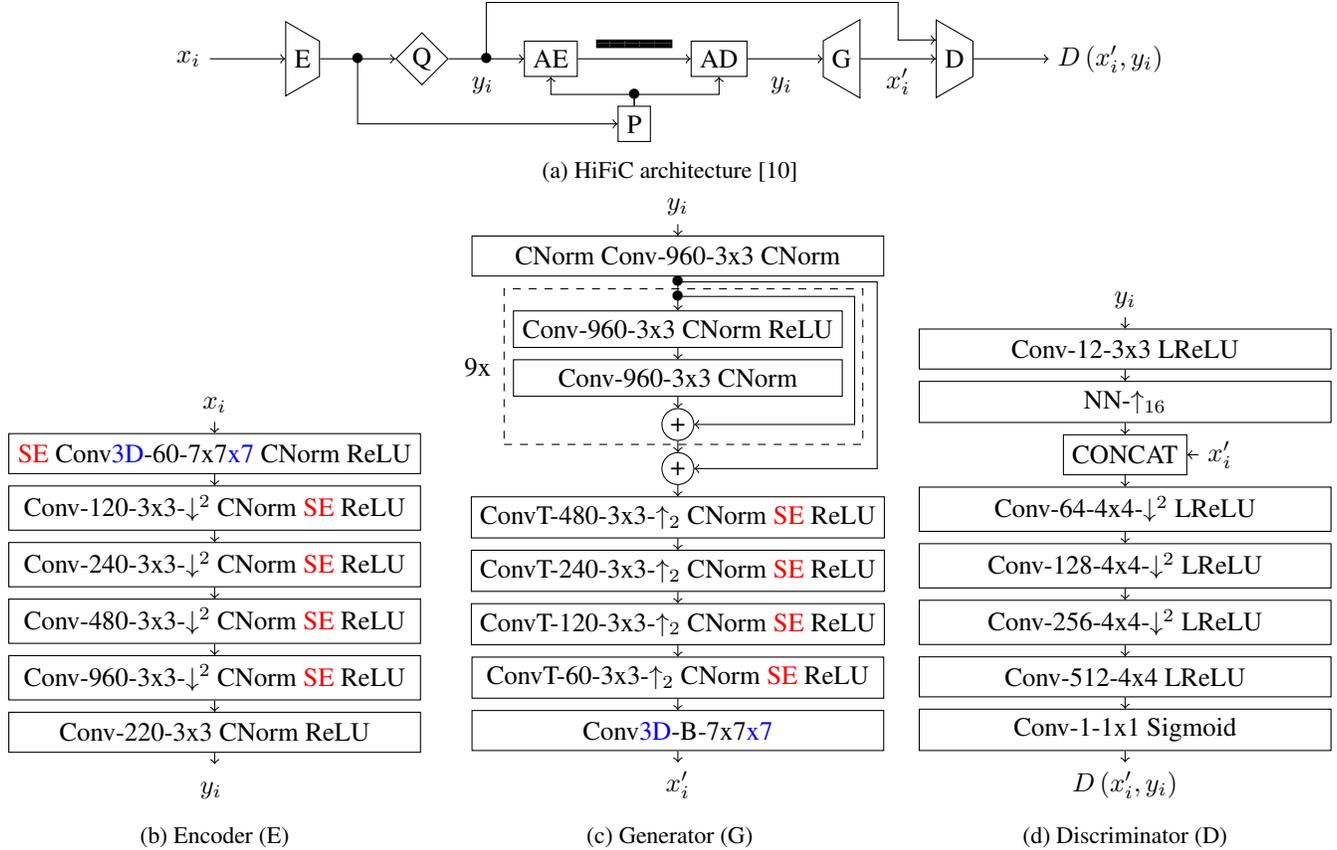
\begin{figure*}[!ht]
    \centering
    \begin{subfigure}{\textwidth}
        \centering\scriptsize
        \begin{tikzpicture}[
            node distance=5mm and 10mm,
        ]
            \node (x) at (0,0) {$x_i$};
            \node[trapezium,draw,shape border rotate=270,right=of x] (e) {E};
            \node[diamond,draw,right=of e,inner sep=1pt] (q) {Q};
            \node[rectangle,draw,right=of q] (ae) {AE};
            \node[rectangle,draw,right=of ae,xshift=5mm] (ad) {AD};
            \node (tmp) at ($(ae.east)!0.5!(ad.west)$) {};
            \node[rectangle,draw,below=of tmp] (p) {P};
            \node[trapezium,draw,shape border rotate=90,right=of ad] (g) {G};
            \node[trapezium,draw,shape border rotate=270,right=of g] (d) {D};
            \node[right=of d] (out) {$D \left( x'_i, y_i \right)$};
            \draw[->] (x) -- (e);
            \draw[->] (e) -- (q);
            \draw[->] (q) -- node[below] {$y_i\phantom{'}$} (ae);
            \draw[->] (ae) -- (ad);
            \draw[->] (ad) -- node[below] {$y_i\phantom{'}$} (g);
            \draw[->] (g) -- node[below] {$x'_i$} (d);
            \draw[->] (d) -- (out);
            \draw[->] (p.north) -- ++(0,0.15) -| (ae);
            \draw[->] (p.north) -- ++(0,0.15) -| (ad);
            \draw[->] ($(q.east)!0.5!(ae.west)$) -- ++(0,0.75) -| ($(g.east)!0.5!(d.west)+(0,0.25)$) -- (d.north west);
            \draw[->] ($(e.east)!0.5!(q.west)$) |- (p);
            \node[rectangle,draw,minimum height=1mm,minimum width=10mm,inner sep=0,outer sep=0, postaction={pattern={Lines[angle=90,distance=2mm,line width=2mm,xshift=-0.8mm,yshift=-0.5mm]}}] (bitstream) at ($(ae.east)!0.5!(ad.west)+(0,0.2)$) {};
            \node at ($(e.east)!0.5!(q.west)$) {$\bullet$};
            \node at ($(p.north)+(0,0.15)$) {$\bullet$};
            \node at ($(q.east)!0.5!(ae.west)$) {$\bullet$};
        \end{tikzpicture}
        \caption{HiFiC architecture \cite{mentzer2020highfidelity}}
        \label{fig:architecture}
    \end{subfigure}
    \begin{subfigure}[b]{0.33\textwidth}
        \centering\scriptsize
        \begin{tikzpicture}[node distance=1.5mm]
            \node (in) {$x_i$};
            \node[block2,below=of x] (1) {{\textbf{\color{red}SE}} Conv{\textit{\color{blue}3D}}-60-7x7{\textit{\color{blue}x7}} CNorm ReLU};
            \node[block2,below=of 1] (2) {Conv-120-3x3-$\downarrow^2$ CNorm {\textbf{\color{red}SE}} ReLU};
            \node[block2,below=of 2] (3) {Conv-240-3x3-$\downarrow^2$ CNorm {\textbf{\color{red}SE}} ReLU};
            \node[block2,below=of 3] (4) {Conv-480-3x3-$\downarrow^2$ CNorm {\textbf{\color{red}SE}} ReLU};
            \node[block2,below=of 4] (5) {Conv-960-3x3-$\downarrow^2$ CNorm {\textbf{\color{red}SE}} ReLU};
            \node[block2,below=of 5] (6) {Conv-220-3x3 CNorm ReLU};
            \node[below=of 6] (out) {$y_i\phantom{'}$};
            \draw[->] (in) -- (1);
            \draw[->] (1) -- (2);
            \draw[->] (2) -- (3);
            \draw[->] (3) -- (4);
            \draw[->] (4) -- (5);
            \draw[->] (5) -- (6);
            \draw[->] (6) -- (out);
        \end{tikzpicture}
        \caption{Encoder (E)}
        \label{fig:e}
    \end{subfigure}
    \hfill
    \begin{subfigure}[b]{0.33\textwidth}
        \centering \scriptsize
        \begin{tikzpicture}[node distance=1.5mm]
            \node (in) {$y_i$};
            \node[block2,below=of in] (1) {CNorm Conv-960-3x3 CNorm};
            \node[block3,below=of 1,align=center,yshift=-3mm] (21) {Conv-960-3x3 CNorm ReLU};
            \node[block3,below=of 21,align=center] (22) {Conv-960-3x3 CNorm};
            \node[circle,draw,below=of 22,inner sep=1.6pt] (23) {+};
            \node[circle,draw,below=of 23,inner sep=1.6pt] (24) {+};
            \node[right=of 22,xshift=-0.5mm,inner sep=0] (2b) {};
            \node[above=of 21,yshift=-1mm,inner sep=0] (2c) {};
            \node[fit=(21)(22)(23)(2b)(2c),draw,yshift=2pt,dashed] (fit) {};
            \node[left=of fit,xshift=3pt] {9x};
            \node[block2,below=of 24] (3) {ConvT-480-3x3-$\uparrow_2$ CNorm {\textbf{\color{red}SE}} ReLU};
            \node[block2,below=of 3] (4) {ConvT-240-3x3-$\uparrow_2$ CNorm {\textbf{\color{red}SE}} ReLU};
            \node[block2,below=of 4] (5) {ConvT-120-3x3-$\uparrow_2$ CNorm {\textbf{\color{red}SE}} ReLU};
            \node[block2,below=of 5] (6) {ConvT-60-3x3-$\uparrow_2$ CNorm {\textbf{\color{red}SE}} ReLU};
            \node[block2,below=of 6] (7) {Conv{\textit{\color{blue}3D}}-B-7x7{\textit{\color{blue}x7}}};
            \node[below=of 7] (out) {$x'_i$};
            \draw[->] (in) -- (1);
            \draw[->] (1) -- (21);
            \draw[->] (21) -- (22);
            \draw[->] (22) -- (23);
            \draw[->] (23) -- (24);
            \draw[->] (24) -- (3);
            \draw[->] (3) -- (4);
            \draw[->] (4) -- (5);
            \draw[->] (5) -- (6);
            \draw[->] (6) -- (7);
            \draw[->] (7) -- (out);
            \draw[->] ($(1)!0.48!(2)$) -- ++(2.65,0) |- (24);
            \draw[->] ($(1)!0.75!(2)$) -- ++(2.35,0) |- (23);
            \node at ($(1)!0.48!(2)$) {$\bullet$};
            \node at ($(1)!0.75!(2)$) {$\bullet$};
        \end{tikzpicture}
        \caption{Generator (G)}
        \label{fig:g}
    \end{subfigure}
    \hfill
    \begin{subfigure}[b]{0.33\textwidth}
        \centering \scriptsize
        \begin{tikzpicture}[node distance=1.5mm]
            \node (in) {$y_i$};
            \node[block2,below=of in] (1) {Conv-12-3x3 LReLU};
            \node[block2,below=of 1] (2) {NN-$\uparrow_{16}$};
            \node[block2,below=of 2,minimum width=0] (3) {CONCAT};
            \node[right=of 3] (x) {$x'_i$};
            \node[block2,below=of 3] (4) {Conv-64-4x4-$\downarrow^2$ LReLU};
            \node[block2,below=of 4] (5) {Conv-128-4x4-$\downarrow^2$ LReLU};
            \node[block2,below=of 5] (6) {Conv-256-4x4-$\downarrow^2$ LReLU};
            \node[block2,below=of 6] (7) {Conv-512-4x4 LReLU};
            \node[block2,below=of 7] (8) {Conv-1-1x1 Sigmoid};
            \node[below=of 8] (out) {$D \left( x'_i, y_i \right)$};
            \draw[->] (in) -- (1);
            \draw[->] (x) -- (3);
            \draw[->] (1) -- (2);
            \draw[->] (2) -- (3);
            \draw[->] (3) -- (4);
            \draw[->] (4) -- (5);
            \draw[->] (5) -- (6);
            \draw[->] (6) -- (7);
            \draw[->] (7) -- (8);
            \draw[->] (8) -- (out);
        \end{tikzpicture}
        \caption{Discriminator (D)}
        \label{fig:d}
    \end{subfigure}
    \caption{\subref{fig:architecture} GAN-based HiFiC model \cite{mentzer2020highfidelity}. \subref{fig:e}, \subref{fig:g} and \subref{fig:d} show the Encoder, Generator and Discriminator architectures of HiFiC$_{opt}$, HiFiC$_{SE}$ with added {\textbf{\color{red}SE}} Blocks and HiFiC$_{3D}$ with two Conv{\textit{\color{blue}3D}} layers, respectively. Q: quantization, P: probability model, AE: arithmetic encoder, AD: arithmetic decoder, Conv{\textit{\color{blue}3D}}(T)-N-KxK({\textit{\color{blue}xK}}): 2D/{\textit{\color{blue}3D}} (transposed) convolution layer with N output channels and kernel size K, $\downarrow^2$/$\uparrow_2$: stride 2, CNorm: channel normalization layer, NN-$\uparrow_{16}$: nearest neighbor upsampling.}
    \label{a}
\end{figure*}

\section{Proposed Models for Spatio-Spectral HSI Compression}
\label{method}

Let $\textbf{X}=\{x_i\}_{i=1}^{H}$ be a set of $H$ uncompressed HSIs, where $x_i$ is the $i^{th}$ image.  The aim of spatio-spectral compression is to produce spatially and spectrally decorrelated latent representations $\textbf{Y}=\{y_i\}_{i=1}^H$ that can be entropy coded and stored with minimal number of bits. At the same time, the learned representations should retain all the necessary information for reconstructing $\textbf{X}'=\{x_i'\}_{i=1}^{H}$ with the least distortion. 

\begin{figure}
  \centering\scriptsize
  \begin{tikzpicture}[node distance=1.5mm]
      \node (input) at (0,0) {$\bullet$};
      \node[block,right=of input] (pool) {Global Pooling};
      \node[block,below=of pool] (fc-mlp) {FC-MLP};
      \node[block,below=of fc-mlp] (relu) {ReLU};
      \node[block,below=of relu] (fc-mlp2) {FC-MLP};
      \node[block,below=of fc-mlp2] (sigmoid) {Sigmoid};
      \node[block,below=of sigmoid] (scale) {Scale};
      \draw (0,0) -- ++(-0.4,0) node[left] {$H \times W \times C$};
      \draw[->] (pool) -- node[right,xshift=12mm] {$1 \times 1 \times C$} (fc-mlp);
      \draw[->] (fc-mlp) -- node[right,xshift=12mm] {$1 \times 1 \times C / r_r$} (relu);
      \draw[->] (relu) -- node[right,xshift=12mm] {$1 \times 1 \times C / r_r$} (fc-mlp2);
      \draw[->] (fc-mlp2) -- node[right,xshift=12mm] {$1 \times 1 \times C$} (sigmoid);
      \draw[->] (sigmoid) -- node[right,xshift=12mm] {$1 \times 1 \times C$} (scale);
      \draw[->] (0,0) |- (scale);
      \draw[->] (0,0) -- (pool);
      \draw[->] (scale.east) -- ++(0.8,0) node[right] {$H \times W \times C$};
  \end{tikzpicture}
  \caption{The basic structure of an SE block \cite{squeeze}.}
  \label{fig:se_layers}
\end{figure}
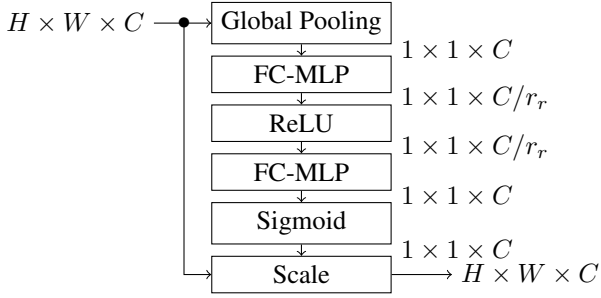

Here, we explain HiFiC$_{SE}$ and HiFiC$_{3D}$ for the spatio-spectral hyperspectral image compression. Both models are defined in the framework of a HiFiC model \cite{mentzer2020highfidelity}. The HiFiC model includes 4 modules: i) encoder $E$; ii) probability model $P$; iii) generator $G$; and iv) discriminator $D$ (see \autoref{a}). The proposed HiFiC$_{SE}$ and HiFiC$_{3D}$ models use the HiFiC as the based model but include SE blocks and 3D convs on the architectures of $E$ and $G$, respectively (see \autoref{fig:e} and \ref{fig:g}). Therefore, we enhance the HiFiC architectures to specifically exploit spatio-spectral redundancies. The HiFiC$_{SE}$ consists of SE blocks prior to the regular 2D convs that enable it to attend to the spectral channel information. The SE block consists of layers of global pooling, Fully Connected Multi-Layer Perceptron (FC-MLP) with ReLU activation, where neuron numbers are decreased by reduction ratio $r_r$, followed by another FC-MLP with Sigmoid activation which outputs weights for each channel (see \autoref{fig:se_layers}). The SE block investigates the relationship between channels by explicitly modelling the inter-dependencies between them. We use $l_1$ regularization at the FC-MLP layers for generating sparse weights indicating the channel importance. These weights are then applied to the feature maps to generate the output of the SE block, which can be fed directly into subsequent layers of the network.
The HiFiC$_{3D}$ is built by replacing the first 2D convs of the Encoder (E) and the last 2D conv of the Generator (G) with 3D convs where the 3D kernels can evaluate spectral redundancies (see \autoref{fig:e} and \ref{fig:g}). It is inspired by the video compression work in \cite{habibian2019video} that uses 3D convs to remove temporal redundancies.

Given an image $x_i \in \textbf{X}$, it is encoded to a spatially and spectrally decorrelated latent representation $y_i$ and then decoded back to $ x_i'$ as:
\begin{equation}
 y_i = Q(E(x_i)), \;
 x_i' = G(y_i)
\end{equation}
$Q$ is a quantizer that is modeled as a rounding operation during inference and straight-through estimator \cite{theis2017lossy} during training. In the bottleneck of the network, the latent $y_i$ is entropy coded into bitstreams using arithmetic encoding (AE) with minimum bitrate $r(y_i)=-\log(P(y_i))$ where $P$ follows the hyper-prior model in \cite{balle2018variational} with side information. The bitstreams are further decoded back to $y_i$ using arithmetic decoding (AD). Finally, a single-scale discriminator $D$ with the conditional information ($y_i$) decides if the reconstructed image is real or generated. Each of the modules $E$, $P$, $G$, and $D$ is parameterized by convolutional neural networks and optimized together to obtain the minimum rate-distortion (RD) trade-off. The loss functions $L$ for optimization of each of the four blocks are as follows:  
\begin{equation}
\label{eq:3}
L_{\scriptscriptstyle E,G,P} = \mathbb{E}_{\scriptscriptstyle x_i \sim P_{\bf{X}}} [\lambda r(y_i) + d(x_i, x_i') - \beta \log D(x_i', y_i)]
\end{equation}
\begin{equation}
L_{\scriptscriptstyle D}  =\mathbb{E}_{\scriptscriptstyle x_i \sim P_{\bf{X}}} [-\log(1-D(x_i', y_i))]+\mathbb{E}_{\scriptscriptstyle x_i \sim P_{\bf{X}}}[-\log D(x_i, y_i)] \\
\end{equation}
where $r(y_i)$ is the rate, $d(x_i,x_i')$ is the distortion loss, $\lambda$ is the hyperparameter controlling the RD trade-off, and $ \log(D(x_i', y_i))$ is the conditional discriminator loss with $\beta$ controlling the discriminator loss effect. 
The distortion loss is modeled as:
\begin{equation}
\label{eq:2}
d(x_i,x_i') = \theta_1 \cdot \textsc{MSE} + \theta_2 \cdot (1 - \textsc{SSIM})+ \theta_3 \cdot \textsc{LPIPS}
\end{equation} 
where $\theta$'s are hyperparameters that control the effect of mean square error (MSE), structural similarity index (SSIM) \cite{wang2004image}, and learned perceptual image patch similarity (LPIPS) \cite{zhang2018perceptual} losses in the total distortion loss calculation. The LPIPS loss mimics the human visual system and measures reconstruction quality in the feature space, while the SSIM loss is added as an improvement to HSI compression as it enhances reconstruction by preserving the structural information relevant to the scene. 
Since $r(y_i)$ is at odds with $d(x_i,x_i')$ and $-\log(D(x_i', y_i))$ in (\ref{eq:3}), controlling the RD trade-off is difficult (as for a fixed $\lambda$, different $\theta$'s and $\beta$ would result in models with different bitrates). Hence, we use target bitrates $r_t$ as in \cite{mentzer2020highfidelity} and control $\lambda$ using $\lambda^{(a)}$ and $\lambda^{(b)}$ with the following rule:
\begin{equation}
  \lambda=\begin{cases}
    \lambda^{(a)}, & \text{if $r(y_i) > r_t$} \\
    \lambda^{(b)}, & \text{otherwise}
  \end{cases}
\end{equation}
 with $\lambda^{(a)} \gg \lambda^{(b)}$ the model learns bitrates closer to $r_t$. By keeping $\lambda^{(b)}$ fixed and changing only $r_t$ and $\lambda^{(a)}$, it is possible to achieve different bitrates. 
 
\section{Dataset Description and Design of Experiments}
We use a hyperspectral dataset that consists of 3,840 image patches with 369 spectral bands (which cover the visible and near-infrared portion of the electromagnetic spectrum in the wavelength range between 400 - 1,000 nm). Each patch is a section of 96 x 96 pixels with a spatial resolution of 28cm. We normalized the data between 0 - 1, which is required for our compression models. The patches were then split in the ratio of 80\%, 10\% and 10\% to form train, validation and test sets, respectively.

All the experiments were carried out using an NVIDIA Tesla V100 GPU with 32 GB of memory. The code for the compression model was implemented with TensorFlow v1.15.2 and built on TensorFlow Compression v1.3 \cite{tfc} and HiFiC \cite{mentzer2020highfidelity}. We trained all the models with Adam Optimizer with a batch size of 8. We use a baseline model trained without GAN ($\beta=0$ in (\ref{eq:3})) as an initializer for our models. For the distortion loss in (\ref{eq:2}) we used $\theta_1 = 0.15 \cdot 2^{-5}$, $\theta_2 = 0.075 \cdot 2^{-3}$ and $\theta_3 =1$. These values were determined through multiple experiments. The discriminator loss effect $\beta$ was set to 0.15, while the reduction ratio $r_r$ in the SE block was selected as 2. As mentioned in Section \ref{method}, different bitrates can be achieved by changing two hyperparameters: i) target rate $r_t$; and ii) $\lambda^{(a)}$. The target rates $r_t$ was varied from [0.2,1] with a step size of 0.2 with its associated $\lambda^{(a)}$ values set accordingly. In total, we configured five settings for each model with different $r_t$ and $\lambda^{(a)}$ combinations (see Table \ref{table:config}).
\begin{table}
  \begin{center}\scriptsize
      \caption{List of target rates $r_t$'s and its associated $\lambda^{(a)}$ values.}
    \label{table:config}
    \begin{tabular}{|l?c|c|c|c|c|} 
     \hline
      $r_t$ & 0.2& 0.4 & 0.6 & 0.8 & 1 \\ 
      \noalign{\hrule height 1.5pt}
      $\lambda^{(a)}$&$2^{1}$&$2^{0}$&$2^{-1}$&$2^{-2}$&$2^{-3}$\\ \hline
      \end{tabular}
  \end{center}
\end{table}
\begin{table}
  \begin{center} \scriptsize
      \caption{Performance analysis with placement of SE and 3D conv blocks in HiFiC$_{SE}$ and HiFiC$_{3D}$, respectively.}
    \label{seblock}
    \begin{tabular}{|l|c|c|} 
     \hline
      \multicolumn{1}{|c|}{SE block placement}& PSNR & Runtime  \\
      & (in dB) & (iters/s)\\
      \hline
     HiFiC$_{SE}$ - $E$: initial layer& 29.80&7.84 \\ \hline
     HiFiC$_{SE}$ - $E$: initial+after Norm&30.12 &7.49\\ \hline
     HiFiC$_{SE}$ - $E$ and $G$: after Norm&30.89&7.01 \\ \hline
     HiFiC$_{3D}$ - $E$ and $G$: all places &30.12&7.31\\\hline
     HiFiC$_{3D}$ - $E$: initial and $G$:  final&29.24 &9.25 \\\hline
      \end{tabular}
  \end{center}
\end{table}
\begin{figure}
  \centering
  \begin{tabular}{ccc}
  \includegraphics[width=0.1\textwidth]{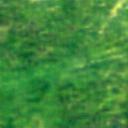}& \hspace{-0.3cm}
  \includegraphics[width=0.1\textwidth]{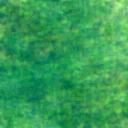}& \hspace{-0.3cm}
  \includegraphics[width=0.1\textwidth]{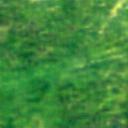}\\
  (a)&(b)&(c)\\
  \end{tabular}
  \caption{(a) Original image; (b) reconstructed image without 3D convs; and (c) reconstructed image with 3D convs.}
  \label{fig:gan_impact}
\end{figure} 
We have compared the proposed spatio-spectral compression models (HiFiC$_{SE}$ and HiFiC$_{3D}$) with: i) the adapted (using the loss in (\ref{eq:2})) and optimized spatial compression HiFiC model (termed as HiFiC$_{opt}$); and ii) the traditional JPEG 2000. We use peak signal-to-noise ratio (PSNR) for quantitative analysis at different compression rates in terms of bitrate measured in bits per pixel (bpp).
\section{Experimental Results}
We initially assess the impact of the choice of placement of SE blocks and 3D convs in the base HiFiC model. The performance is evaluated in terms of: i) the reconstruction quality measured with average PSNR and ii) runtime calculated as the average number of iterations completed per second (denoted as iters/s) to decide the best placement. Note that the higher the value of iters/s, the faster the model. The obtained metric by placing the SE and 3D blocks at different layers in the model is provided in Table \ref{seblock}. For this, the SE blocks were placed in the initial layer of $E$. Then, we placed it in the initial layer and after every normalization layer in $E$, and finally, we extended SE blocks to cover all layers in $E$ and $G$ after every normalization layer. We observe that the best PSNR with comparable runtime to other placements is achieved, when the SE block is incorporated in both $E$ and $G$ after the normalization layer.
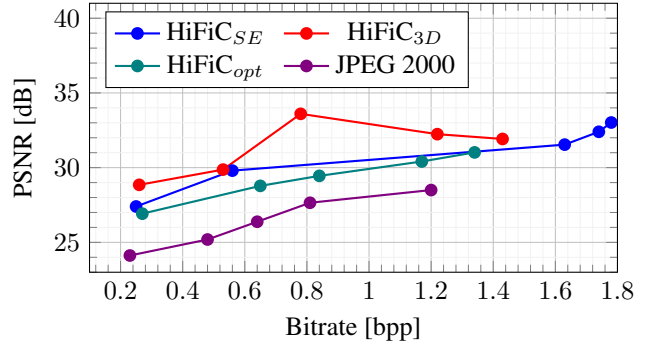
\begin{figure}
    \centering\scriptsize
    \begin{tikzpicture}
        \begin{axis}[
            xlabel={Bitrate [bpp]},
            ylabel={PSNR [dB]},
            xmin=0.1,xmax=1.8,
            ymin=23,ymax=41,
            legend pos=north west,
            width=\linewidth,
            height=0.6\linewidth,
            legend columns=2, 
            legend style={
                /tikz/column 2/.style={
                    column sep=5pt,
                },
            },
        ]
    	    \addplot[color=blue,mark=*] coordinates {
                (0.25,27.40)
        		(0.56,29.80)
                (1.63,31.54)
            	(1.74,32.40)
            	(1.78,33.02)

        	};
             \addplot[color=red,mark=diamond*] coordinates {
    		    (0.26,28.85)
    		    (0.53,29.87)
        		(0.78,33.60)
        		(1.22,32.24)
        		(1.43,31.92)
            };
            \addplot[color=teal,mark=triangle*] coordinates {
                (0.27,26.92)
                (0.65,28.78)
                (0.84,29.45)
                (1.17,30.41)
                (1.34,31.02)
            };
            \addplot[color=violet,mark=square*] coordinates {
        		(0.23,24.12)
        		(0.48,25.19)
            	(0.64,26.38)
            	(0.81,27.65)
            	(1.2,28.5)
            };
            \legend{HiFiC$_{SE}$,HiFiC$_{3D}$, HiFiC$_{opt}$, JPEG 2000}
    	\end{axis}
    \end{tikzpicture}
    \caption{Test set rate-distortion performance of our proposed HiFiC$_{SE}$ and HiFiC$_{3D}$ compared to HiFiC$_{opt}$ and JPEG2000.}
    \label{fig:graph}
\end{figure}
\begin{figure}[htb]
  \centering
  \begin{tabular}{cccc}
  \includegraphics[width=0.1\textwidth]{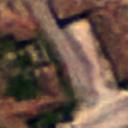}& \hspace{-0.3cm}
  \includegraphics[width=0.1\textwidth]{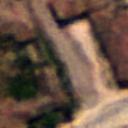}& \hspace{-0.3cm}
  \includegraphics[width=0.1\textwidth]{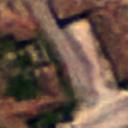}&\hspace{-0.3cm}
  \includegraphics[width=0.1\textwidth]{images/reconstructions/output_2.jpeg}\\
  &\small{(29.17, 0.82)}&\small{(29.34, 1.66)}&\small{(30.33, 0.55)}\\
   (a)&(b)&(c)&(d)\\
  \end{tabular}
  \caption{(a) Original. Reconstructed with (b) HiFiC$_{opt}$, (c) HiFiC$_{SE}$ and (d) HiFiC$_{3D}$. The PSNR and achieved bpp are shown below each image as (PSNR in dB, bpp). }
  \label{viscomp}
\end{figure}
We also analyze the importance of replacing 3D conv layers in the GANs for reconstruction. Figure \ref{fig:gan_impact} shows the reconstructed images with HiFiC$_{SE}$ and HiFiC$_{3D}$ using the GAN. The reconstruction without 3D conv layers in \autoref{fig:gan_impact}(b) fails to generate the spatial information characterized by the texture and geometrical characteristics of the objects within images. But with the integrated 3D conv layers in \autoref{fig:gan_impact}(c) the perceptual quality is improved and the texture content is better represented. In addition, high-quality compressed images are obtained with reduced perceptible artifacts. This shows the importance of the 3D convs in GAN-based models for HSI compression. \autoref{fig:graph} shows the rate-distortion curves for the spatio-spectral compression methods HiFiC$_{SE}$ and HiFiC$_{3D}$ compared with the optimised spatial compression method HiFiC$_{opt}$ and the traditional JPEG 2000 on the test set. We can observe that all the end-to-end deep compression methods outperform JPEG 2000 at these low bitrates consistently. Amongst the spatial and spatio-spectral compression methods, we can observe that the HiFiC$_{3D}$ performs slightly better than HiFiC$_{SE}$ and HiFiC$_{opt}$ throughout most bitrates.
We observe that the proposed model HiFiC$_{3D}$ achieves significantly higher PSNR values compared to other methods. The qualitative results of reconstruction at $r_t=0.6$ with the proposed spatio-spectral compression models in comparison with HiFiC$_{opt}$ are also shown in \autoref{viscomp}. The qualitative results look similar with spatial and spatio-spectral compression methods. However, on closer inspection, we notice a slight improvement in edge details and sharpness that results in better PSNR at lower bpp with the spatio-spectral compression models HiFiC$_{SE}$ and/or HiFiC$_{3D}$ when compared to HiFiC$_{opt}$. 

\section{Conclusion}
In this paper, we have presented two High Fidelity Compression (HiFiC)-based models for spatio-spectral compression of hyperspectral images (HSIs). The first model is the HiFiC$_{SE}$ that contains Squeeze and Excitation (SE) blocks, while the second model is HiFiC$_{3D}$ that includes 3D convolutions. Experimental results show that the proposed models remarkably outperform JPEG 2000 by reducing the redundancy in latent representations. In addition, compared to the spatial compression model HiFiC$_{opt}$, the proposed  HiFiC$_{SE}$ and  HiFiC$_{3D}$ provide higher PSNR values for the same bitrate producing perceptually better reconstructions with richer details.

\bibliographystyle{IEEEtran}
\bibliography{refs.bib}

\end{document}